\documentclass[letterpaper, 10 pt, conference]{ieeeconf}  % Comment this line out if you need a4paper

\IEEEoverridecommandlockouts                              % This command is only needed if 
\title{\LARGE \bf
Applications of Uncalibrated Image Based Visual Servoing\newline in Micro- and Macroscale Robotics
}

\author{Yifan Yin, Yutai Wang, Yunpu Zhang,\\Russell H. Taylor, \emph{Life Fellow, IEEE} and Balazs P. Vagvolgyi, \emph{Member, IEEE}%
\thanks{*This work was supported in part by NIH grant 2R44AI134500-04A1 in collaboration with Sanaria, Inc. Rockville, MD, USA and in part by Johns Hopkins University internal funds.}
\thanks{All authors are with The Laboratory for Computational Sensing and Robotics (LCSR) at the Johns Hopkins University, Baltimore MD, USA.\newline
\noindent
{\tt\footnotesize \hspace*{0.3cm}\{yyin34, ywang790, yzhan625, rht, balazs\}@jhu.edu}}%
}

\usepackage{gensymb}
\usepackage{subfig}
\usepackage{diagbox}
\usepackage{amsfonts}
\usepackage{newtxmath}
\usepackage{adjustbox}
\usepackage{balance}
\usepackage{graphicx}
\graphicspath{{figs/}}
\usepackage{floatrow}
\UseRawInputEncoding

\begin{document}
\maketitle
\thispagestyle{empty}
\pagestyle{empty}

%%%%%%%%%%%%%%%%%%%%%%%%%%%%%%%%%%%%%%%%%%%%%%%%%%%%%%%%%%%%%%%%%%%%%%%%%%%%%%%%
\begin{abstract}

We present a robust markerless image based visual servoing method that enables precision robot control without hand-eye and camera calibrations in 1, 3, and 5 degrees-of-freedom. The system uses two cameras for observing the workspace and a combination of classical image processing algorithms and deep learning based methods to detect features on camera images. The only restriction on the placement of the two cameras is that relevant image features must be visible in both views. The system enables precise robot-tool to workspace interactions even when the physical setup is disturbed, for example if cameras are moved or the workspace shifts during manipulation. The usefulness of the visual servoing method is demonstrated and evaluated in two applications: in the calibration of a micro-robotic system that dissects mosquitoes for the automated production of a malaria vaccine, and a macro-scale manipulation system for fastening screws using a UR10 robot. Evaluation results indicate that our image based visual servoing method achieves human-like manipulation accuracy in challenging setups even without camera calibration.

\end{abstract}

%%%%%%%%%%%%%%%%%%%%%%%%%%%%%%%%%%%%%%%%%%%%%%%%%%%%%%%%%%%%%%%%%%%%%%%%%%%%%%%%
% \input{introduction}
\section{Introduction}
\label{sec:introduction}

Precise unsupervised robotic manipulation requires accurate knowledge of the spatial relationship between the robot, the tools that are mounted on it and relevant objects in the workspace. In many applications the robot controller relies on known geometrical models or constraints to calculate its actions in an open-loop system. In other applications, various types of sensors provide information to plan the robot's actions or guide its movements using closed-loop control. These sensors, such as cameras, are usually calibrated both optically and with respect to the workspace so that the results of robot motions relative to other objects can be planned and predicted in three-dimensional (3D) space.

%In most applications, robot motion planning is done either in joint space or in Cartesian space. For non-Cartesian robots with multiple degrees-of-freedom, joint space control is typically used in situations where the robot is not expected to interact with other objects in its workspace, as these movements may result in unintuitive actions that may otherwise result in collisions. For example, joint space control may be used for adjusting the pose of a robot arm to move far from singularities between periods of Cartesian space control.

Cameras serve as excellent sensors for motion tracking due to their large field of view and high resolution. Measuring their optical properties using a calibration process enables precise 3D geometric calculations based on features located in 2D camera images. The calculation of the position and orientation of an object visible in a single calibrated camera's view with respect to the camera -- a method called pose estimation -- has been widely demonstrated. Perspective-N-Point (PnP) is the name of a family of computational algorithms~\cite{lu2018review}\cite{lepetit2009ep}\cite{xu2008general} that calculate accurate object pose based on the precise image coordinates of four or more visible point-like object features with known geometry. As PnP methods require known point cloud geometry, they are ideal for tracking optical markers, such as AR tags~\cite{garrido2016generation}\cite{romero2018speeded} and 3D rigid bodies constructed from point-like markers (e.g. light emitting diodes, retroreflective spheres)~\cite{vanzella2019passive}\cite{faessler2014monocular}\cite{vagvolgyi-tracker}.

\begin{figure}[!ht]
  \centering
  \vspace{7pt}
  \includegraphics[width=1.0\textwidth]{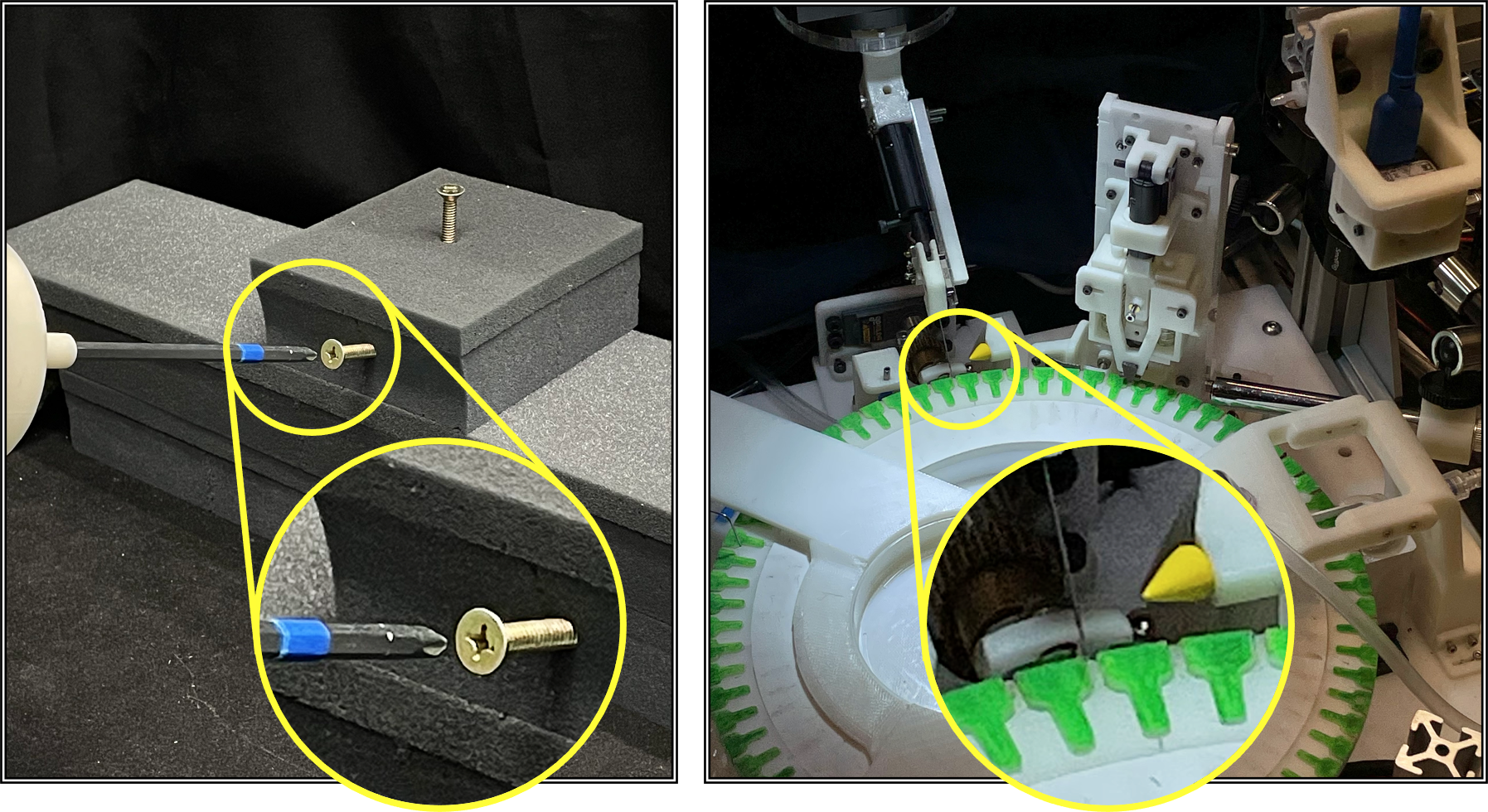}
  \caption{(Left) Macroscale task of aligning screw driver with screw. (Right) Automated calibration of a robotic mosquito salivary gland extraction system that uses a micro-forceps for manipulating mosquitoes.}
  \label{fig:front-page-pic}
%  \vspace{-3pt}
\end{figure}

For objects that lack adequate number of accurately identifiable point-like features, machine learning-based (ML) methods have been developed that use deep neural networks to infer position and orientation parameters directly from images~\cite{PoseCNN}. While ML based methods can be adapted for a substantially wider range of object appearances, their accuracy lags significantly compared to PnP methods in 3 degrees-of-freedom (DoF) orientation detection~\cite{zhou2019continuity}\cite{liang2021manufacturing}. There also exist a class of hybrid methods that use ML-based methods to detect point-like features on objects from multiple  camera poses to estimate the 3D positions of these features~\cite{nath2019using}\cite{gunel2019deepfly3d}\cite{li2020}.

While single camera (monocular) pose estimation methods are quite accurate, they are very sensitive to measurement noise~\cite{ferraz2014leveraging}, therefore determining object pose based on multiple camera views is the gold standard of robust high precision 3D pose tracking. Methods based on multiple distinct camera views have two main advantages: they are less sensitive to occlusions and object coordinates can be estimated to high accuracy using triangulation even in the presence of a moderate amount of image noise. To enable 3D triangulation, the relative 3D poses of cameras (extrinsic parameters) need to be known, which is a drawback of multi-camera systems. This shortcoming is usually addressed by either mounting the pre-calibrated multi-camera rig on a rigid physical frame or by performing extrinsic camera calibration every time cameras are moved with respect to each other. Most research, industrial and medical optical tracking devices use two or more cameras for 3D object tracking~\cite{optitrack}\cite{naheem2022}\cite{Wang2016RoboticSW}.

In all of the previously mentioned 3D pose estimation methods, the object pose is calculated in 3D Cartesian space. When a robotic system needs to interact with its workspace in a precise manner, motion planning is usually done in Cartesian space~\cite{vs_tutorial}\cite{gorner2019}. Also, Cartesian space is the favored spatial representation of most object modeling and simulation systems, either in two- or three-dimensions. Since relating poses in different Cartesian representations in straightforward, having robot motions planned in Cartesian space makes it easy to relate these actions to other objects in the workspace.
%While robot control in 2D Cartesian space is certainly routinely done, its use is limited to applications where the objects that the robot needs to manipulate are arranged along a flat surface and actions along the third dimension are not part of trajectory, such as actuating motions (for example pushing a button). In this work, we are focusing primarily on robot control in 3D Cartesian space.

It is also possible to plan robot actions directly in image space~\cite{janabi2010comparison}\cite{Yang2023}\cite{TSUCHIDA202176}\cite{wang2016adaptive}\cite{feemster2020autonomous}\cite{gong2017uncalibrated}. As we pointed out in the previous paragraphs, an object's 3D pose can be often inferred even from a single 2D image and having multiple camera views allows better accuracy; the inverse of this, translation from 3D space to 2D image space is simple, and in fact a fundamental task of 3D computer graphics. This isomorphic property of projective transformations enables planning motion trajectories based on the appearance of objects in the camera view.

To take advantage of the isomorphism one can simulate the appearance of relevant visual features of objects in the camera view during robot motion and then servo the robot along a trajectory that minimizes some error metric between the feature positions in simulation and on the camera images. While generating realistic simulated images is possible in situations where camera calibrations and object models are known, simulations may not always need to be very realistic. In cases where the objects that need to interact with each other are all visible in the cameras and the relevant aspects of their spatial relationship can be derived in image space, only the metrics that express these relevant relationships need to be calculated.

For example, in the classic peg-and-hole task it is enough to know that (1) the axis of the peg matches the orientation of the minor axis of the ellipse representing the hole, (2) the axis of the peg points to the center of the hole, and (3) the peg moves in the direction of the hole during insertion. All these properties are defined in 2D image space, although a single camera view is not enough to servo the robot in 3D space unless further, more complex properties are taken into account. However, adding a second camera view from a substantially different view angle and extracting image features (1-3) from the second image as well enables highly accurate visual servoing for peg insertion. Interestingly, this method is so robust, that we can even omit intrinsic and extrinsic camera calibrations. It can be demonstrated that the same visual servoing method will work even if the cameras or the objects are not static but moving in an unknown way, as long a the objects stay visible and detectable in both camera views.

The root cause of this level of robustness is that the used 2D properties (1-3) are all invariant to image transformations that result from a 3D pose change followed by perspective projection: rigid 3D transformations and perspective projection preserve collinearity.

In the followings we demonstrate the usefulness of this visual servoing control scheme in two applications: one is a system with a visually guided 3-DoF Cartesian robotic micro-manipulator~\cite{mosquito2021icra}\cite{mosquito2021case}, where the calibration of the workspace is performed using visual servoing, the second example is a macro-scale manipulation task in 5-DoF using a UR10 robot (Universal Robots Inc., Odense, Denmark) in which the goal is to align a robot-held screw driver with screws in the workspace. Our experiments compared the performance and accuracy of visual servoing based control to manual control by an operator.

\section{Methods}
\label{sec:methods}
\subsection{Concepts}
Our investigation focuses on one type of image based visual servoing (IBVS) method in which cameras are mounted somewhere adjacent to the workspace independently from the robot, the so called “hand-to-eye” configuration. Consider a robot with tool coordinates $\boldsymbol{r}$ in some m-dimensional parameter space, and $\boldsymbol{\dot{r}}$ represents the time rate of change of those coordinates. In case of a linear robot, $\boldsymbol{r}$ can be the position vector of the tool in Cartesian space; in case of a robotic manipulator, $\boldsymbol{r}$ can be any parameterized representation of the end effector pose. The vector $\boldsymbol{f}\in \mathbb{R}^k$ represents the image features, where $k$ is the number of features. In the general case, the relationship between the visual features and the robot's motion can be described using the image Jacobian $J$, which is defined as
\begin{equation}
    \dot{\boldsymbol{f}}=J(\boldsymbol{r})\dot{\boldsymbol{r}}
\end{equation}
where $J \in \mathbb{R}^{k\times m}$, and
\begin{equation}
J(\boldsymbol{r})=\left[\frac{\partial \boldsymbol{f}}{\partial \boldsymbol{r}}\right]=\left[\begin{array}{ccc}
\frac{\partial f_1(\boldsymbol{r})}{\partial r_1} & \cdots & \frac{\partial f_1(\boldsymbol{r})}{\partial r_m} \\
\vdots & & \vdots \\
\frac{\partial f_k(\boldsymbol{r})}{\partial r_1} & \cdots & \frac{\partial f_k(\boldsymbol{r})}{\partial r_m}
\end{array}\right]
\end{equation}

\subsection{Update of the image Jacobian}
In the absence of camera and hand-eye calibrations we are not able to initialize the image Jacobian (also called the interaction matrix in the literature) from a camera image; furthermore, for safety reasons we do not have the opportunity to learn the Jacobian during manipulation using machine learning techniques (such as \cite{gaskett2000jacobian}\cite{shi2020jacobian}). Instead, we choose to evaluate the inital Jacobian value $J_{0}$ using finite difference approximation
\begin{equation}
    {J_0}_{ij} = \frac{f_i(x_1, ..., x_j + h, ..., x_n) - f_i(x_1, ..., x_j - h, ..., x_n)}{2h}
\end{equation}
This initial value is only valid locally near the position where it gets calculated. To update the image Jacobian while the robot is in motion, an optimization scheme is constructed as follows
\begin{equation}
\hat{J}\left(\boldsymbol{r}_{t+1}\right)=J_{t+1}=\underset{J_t}{\operatorname{argmin}}(J \cdot \Delta \boldsymbol{r}-\Delta \boldsymbol{f})
\end{equation}
where $\Delta \boldsymbol{r}$ is the change in tool coordinates during an update period and $\Delta \boldsymbol{f}$ is the corresponding change in feature space. This optimization scheme takes the Jacobian approximation at time $t$ as the initial value then finds the new Jacobian that minimizes the reprojection error for time $t+1$. These iterative corrections enable the system to keep the image Jacobian up-to-date by making use of the most recent robot sensing data and the previous time rate of change. In fact, the updated Jacobian is an approximation of the local Jacobian at some point along the robot's trajectory since the last update. In our visual servoing software, we use the Levenberg-Marquardt (LM) algorithm to solve the optimization problem. Although it is an inherently under-determined optimization problem, the robustness of the method is supported by incorporating the Jacobian from the previous iteration as the initial condition, and further improvements are possible by increasing the sampling frequency, or introducing regularization terms in the objective function.

\subsection{Control laws}

The control law of this visual servoing method is given by
\begin{equation}
\boldsymbol{f}^*-\boldsymbol{f}_t=\hat{J}\left(\boldsymbol{r}_t\right) \Delta \boldsymbol{r}
\end{equation}
where $\boldsymbol{f}^*$ is the target feature vector and $\Delta \boldsymbol{r}$ is the least squares solution of the system. In this work, we consider merely the cases where the number of features is greater than the number of robot coordinates. Thus, the velocity of the robot tool can be written as  
\begin{equation}
\Delta \boldsymbol{r} = K* J^{+} \cdot (\boldsymbol{f}^*-\boldsymbol{f}_t)
\end{equation}
where $K$ is the gain of the resolved-rate control and $J^{+}$ is the pseudo-inverse of the image Jacobian.

\subsection{Features of visual servoing}
\label{features-of-visual-servoing}

In our method, the position and orientation of the robot are controlled separately using distinct visual features. For the position servo, point features are used for both robot tool tracking and target localization, represented as image coordinates. The feature vector is given by
\begin{equation}
\boldsymbol{f}=[u_1, v_1, \ ... \ u_n, v_n]^T
\end{equation}
where $(u_i, v_i)$ are the image coordinates of the same point feature in the $i^{th}$ camera views.

To generate features for the orientation servo, first consider a rigid body in the camera view, which has a rigid frame attached to it. Each of the three (3D unit-length) basis vectors of this frame will have its corresponding projection (2D vector) in the image plane, which can be represented by a single parameter, the image angle $\theta$. In the robot control process, the goal of the orientation servo is to align the projected basis vectors of the tool frame with the projected basis vectors of the target object frame. Since the third axis of a frame can be determined by the cross product (in correct order) of the other two, aligning two vectors in the image will be sufficient. 

Although it is possible to use angular distances in image space as the feature for orientation servoing, it is challenging to calculate image Jacobian by using angle as the rotation parameter. In this work, an energy term is proposed in order to solve the issue of circular angle space. The energy term takes the following form 
\begin{equation}
E = \cos(\theta^{*}-\theta_t)-1
\end{equation}
where $\theta_t$ is the current image angle of a vector attached to the robot tool, $\theta^{*}$ is the target image angle. This energy term will go to zero if the two image vectors are perfectly aligned and its norm will be greatest if they have the opposite direction. Thus, the goal term in the orientation servo is always a zero vector. The final feature vector for the orientation can be written as

\begin{equation}
\boldsymbol{f}=[E_{u_1}, E_{v_1}, \ ... \ E_{u_n}, E_{v_n}]^T
\end{equation}
where $E_{u_i}$ is the energy term of the first image angle for the $i^{th}$ camera, $E_{v_i}$ is that of the second image angle.

\subsection{Detection of image features}

The feature vector described in Sec.~\ref{features-of-visual-servoing} is derived from point-like image features in our implementation. For the real-time detection of these features, we employed classical image processing (IP) algorithms and high-performance machine learning (ML) methods.

\subsubsection{ML}
\label{feature-detector-ml}
We used the object detector library YOLOv5~\cite{yolov5} for detecting image features with variable appearances at different view angles and illumination. The detected features are represented by a rectangle, the center of which was used as the point-like feature in our control method. YOLOv5 is an object detection algorithm known for its speed and accuracy in real-time detection. It uses a single convolutional neural network to directly predict object class probabilities and bounding box coordinates, allowing it to process images quickly.
We used data augmentation and domain randomization techniques to achieve robust feature detection, i.e. translation, scaling, rotation, lighting change, defocus blur, and background replacement.
%A summary of the training information of the network for both tasks can be found in Table X. 
% (put the table here) (Balazs: what is in the table?)

%YOLOv5 also employs a novel training strategy involving data augmentation techniques that improve model generalization and accuracy. It has achieved state-of-the-art performance on multiple object detection benchmarks while maintaining real-time inference speeds, making it a popular choice for real-world applications.

\subsubsection{IP}
\label{feature-detector-ip}
Our experimental apparatuses also include physical targets with appearances that are easily detectable by simple classical image processing approaches.
One of these feature detectors is capable of detecting patches of certain color on the image and calculate the centroid of the largest such patch, which in turn represents the point-like image feature corresponding to the physical target of that color.
Another feature detector is an enhanced version of the color patch detector, which first detects the largest color patch on the image, then finds its contours and fits straight lines along the edges using Hough transform to enable the precise calculation of the corner points of such objects.

\begin{figure*}[t]
  \centering
  \includegraphics[width=1.0\textwidth]{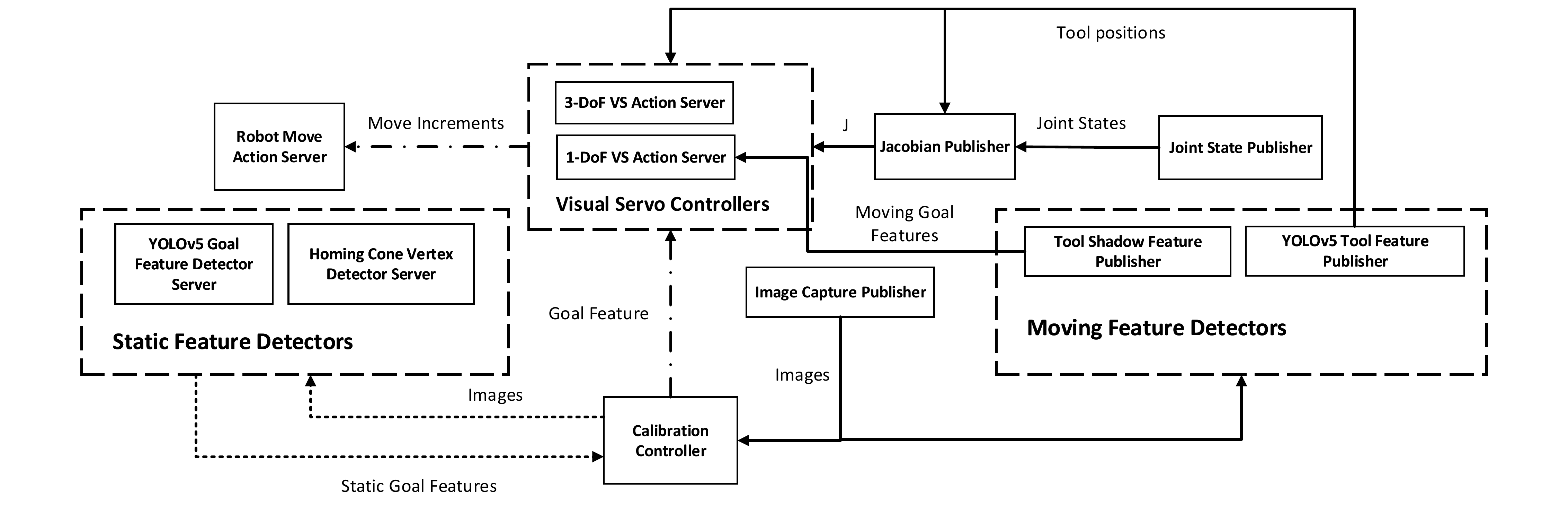}
  \vspace{-5pt}
  \caption{Block diagram illustrating the ROS nodes and their connections while performing visual servo control for the calibration of the mosquito salivary gland extraction system. Dashed rectangles are logical blocks. Solid arrows represent topics, pointing from publisher to subscriber; dotted arrows are service calls; dotdashed arrows stand for action calls.}
  \label{fig:system-block-diagram}
%  \vspace{0pt}
\end{figure*}

\section{System Design}
\label{sec:system}

Our IBVS method is implemented using the Robot Operating System (ROS)~\cite{ros_website}, which allows for efficient communication and control of components involved in the process. The software components (ROS nodes) and data flow within the system are illustrated in the block diagram in Fig.~\ref{fig:system-block-diagram}.

The main control hub of the system is a Task Controller (e.g. Calibration Controller) node, which select which visual servoing task to execute and directs data flow accordingly. The Joint State Publisher for the robot, the Image Capture Publisher nodes, and the two Moving Feature Detectors (Tool Shadow Feature Publisher and YOLOv5 Tool Feature Publisher) run continuously and publish their most recent data in ROS topics. The two Static Feature Detectors (YOLOv5 Goal Feature Detector Server and Homing Cone Vertex Detector Server) are called when needed by the Task Controller. The Jacobian Publisher node, which subscribes to the joint states and the tool detector features, updates the Jacobian for every video frame and publishes the updated Jacobian in a ROS topic.

The Visual Servo Controllers are called by the Task Controller when a servoing task needs to be executed. These nodes subscribe to the outputs of Moving Feature Detectors and receive the static features from the Task Controller, and their job is to run the visual servoing control loop and send the robot motion commands to the Robot Move Action Server. The Visual Servo Controllers are also responsible for determining when the tool reached the goal position, finish the servoing operation and notify the Task Controller of the results.

\section{Experimental Setup}
\label{sec:experiments}

\subsection{\textbf{EXP-1:} Calibration of mosquito dissector}	
\label{sec:exp1}

The primary motivation for our visual servoing method was the need to create an autonomous calibration technique for a micro-scale robotic system that was developed for automated mosquito salivary gland extraction~\cite{mosquito2021icra}\cite{mosquito2021case}. The system aims to accelerate the production of a malaria vaccine that requires mosquito salivary glands for manufacturing.
Our mosquito dissection system uses a micro-forceps mounted on a 3-DoF (3D position) Cartesian stage and a motorized blade to decapitate mosquitoes, then it squeezes the salivary glands out of the thorax and collects the specimen. The robotic system features a large number of replaceable and moving parts, all of which need to be aligned to each other at a very high precision. After installation, part replacement or servicing, the system needs to be routinely re-calibrated which is done under the guidance of high-magnification cameras. Prior to having an automated calibration method, manual calibration often took over 30 minutes for a trained operator. Due to compact size of the apparatus, camera intrinsic and extrinsic calibrations are not possible, therefore the spatial relationship of the robot with respect to other objects can only be established by getting infinitesimally close to reference points with the micro-forceps.

%The waypoint calibration process involves using the calibration-free visual servoing method to automatically calibrate the robot's encoder counts for the critical waypoints during mosquito dissection.
The calibration setup consists of two camera views: overhead view and side view. Specific points of interest are detected on these views, which the robot then approaches with the robotic micro-forceps under 3-DoF IBVS control. A YOLOv5-based network was trained to track features in real-time on the micro-forceps during all visual servoing tasks throughout the entire calibration process (see Sec.~\ref{feature-detector-ml}). The calibration steps are the following:

\subsubsection{Calibration of robot home position}
\begin{figure}[h]
  \centering
  \includegraphics[width=0.7\textwidth]{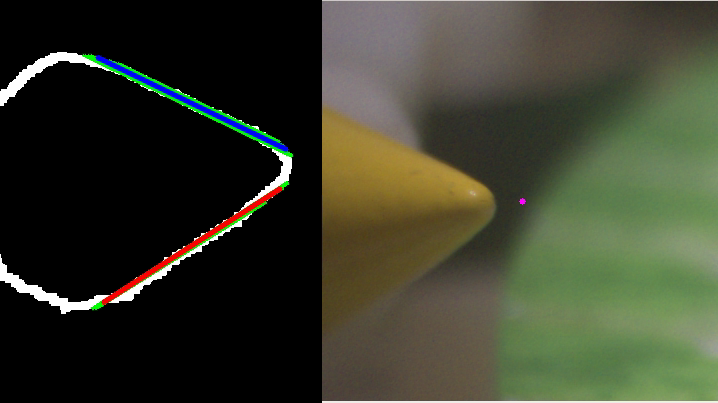}
  \caption{\label{fig:Res} Homing cone vertex detection using color segmentation and Hough transform. The small purple dot shows the location of the detected vertex.}
  \label{fig:conevertex}
  \vspace{-5pt}
\end{figure}

(A) Detection of the vertex of the yellow conical homing marker in the camera views using an IP-based yellow cone vertex detector (see Sec.~\ref{feature-detector-ip} and Fig.~\ref{fig:conevertex}). The physical conical marker is not a perfect cone: its vertex is missing due to the rounding of the tip. The goal of visual servoing is to move the micro-forceps to the geometrical vertex of the cone, a point that is about a 0.5\,mm distant from the rounded tip, therefore the sensitive micro-forceps is in no danger of touching the cone upon approach. (B) Visual servoing of the robotic micro-forceps (gripper closed) to the cone vertex. (C) Storage of robot joint positions.

\subsubsection{Calibration of cutting blade center}
(A) Detection of the center of the cutting blade in both camera views using a YOLOv5 detector (see Sec.~\ref{feature-detector-ml}). (B) Visual servoing the robotic micro-forceps to the cutting blade center. (C) Storage of the robot joint positions.

\subsubsection{Calibration of mosquito tray surface}
(A) Moving the robot to a position above the mosquito tray. (B) Performing 1-DoF visual servoing to approach the surface of the tray. The input feature for servo control is the distance between the detected tip of the micro-forceps and the detected location of its shadow on the tray as measured on the side view camera's image. Stopping when the tip and its shadow coincides. (C) Storage of the position of the robot's Z joint.

\subsection{\textbf{EXP-2:} Robotic screw driver}
\label{sec:exp2}

\begin{figure}[htbp]
  \centering
  \includegraphics[width=0.9\linewidth]{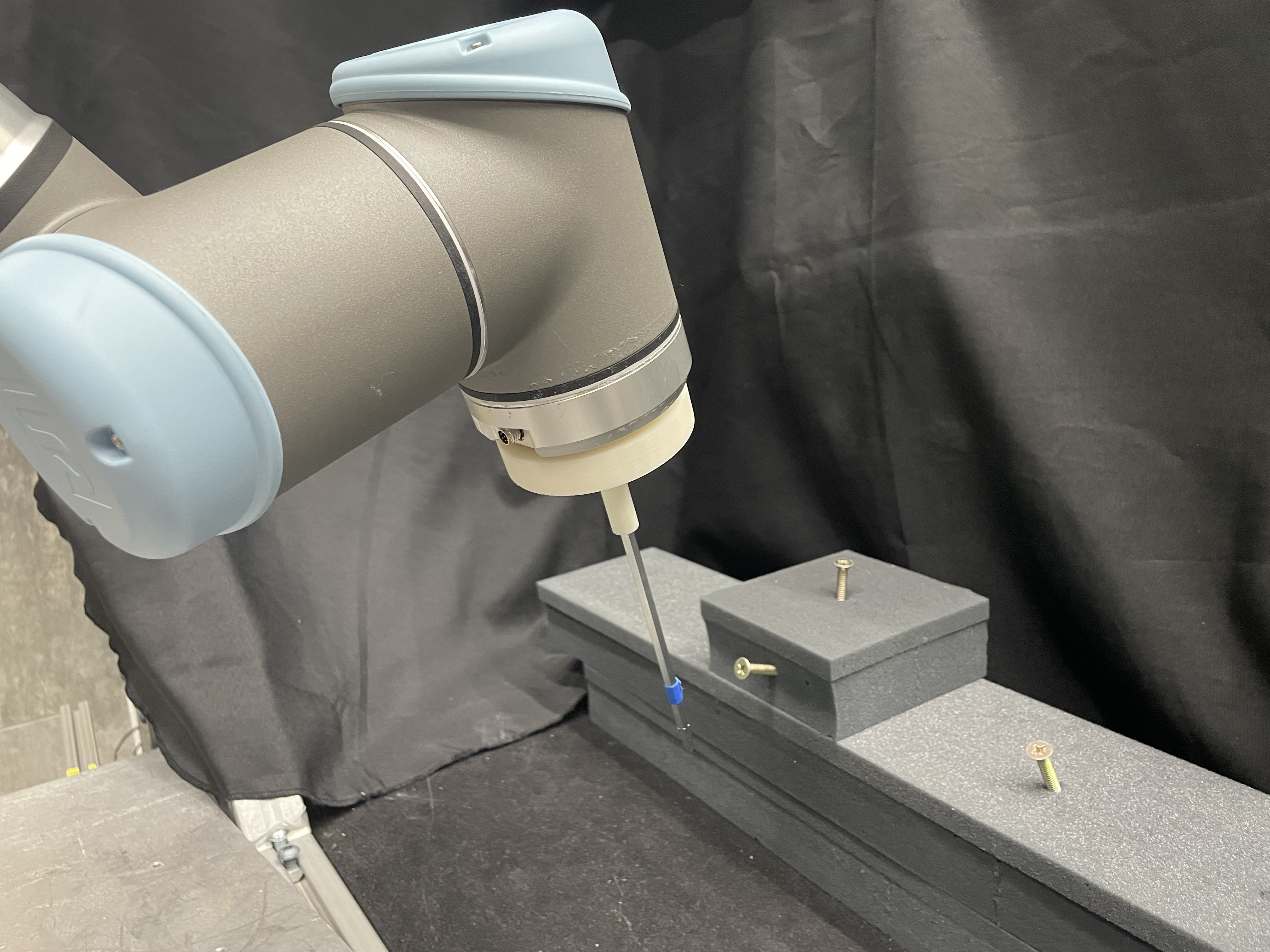}
  \caption{The workspace of \textbf{EPX-2} (robotic screw driver experiment). The screw driver is mounted on the UR10's end-effector and 3 screws are staged on the platform.}
  \label{fig:workspace}
  \vspace{-5pt}
\end{figure}

This experiment aims to demonstrate the applicability of our IBVS method in a 5-DoF (3-DoF position + 2-DoF orientation) macro-scale task using a UR10 robot arm. The task consists of aligning a robot-held screw driver with three screws of different positions and orientations in the workspace, then turning the screws using the screw driver, as shown in Fig.~\ref{fig:workspace}. Our goal is to demonstrate the robustness and accuracy of our calibration-free visual servoing method in a real-world scenario.

\subsubsection{System setup}
The experimental setup consists of a UR10 robotic arm equipped with a custom end-effector to hold a screwdriver. Two cameras are placed in the workspace, observing the screwdriver and the screws from different angles. The placement of the cameras must ensure that the relevant image features are visible in both views.

\subsubsection{Image feature detectors}
The screw driver is represented by two features in image space: (1) a point feature for the tip of the screw driver, detected using a YOLOv5 detector (see Sec.~\ref{feature-detector-ml}) that was trained on thousands of images generated using domain randomization and data augmentation, and (2) an angle feature for the direction of the screw driver shaft in image space, which is calculated by locating a blue marker on the shaft using IP-based color segmentation (see Sec.~\ref{feature-detector-ip}). Screws are described by two image point features, one for the head and one for the point on the shank that is level with the surface of the platform, both of which are detected by a YOLOv5 detector trained on hundreds of images.
%(A table with training information is provided below. Balazs: do we need it? is there room???)

% Add the table here

\subsubsection{Evaluation method}
Visual servoing performance is compared to manual alignment results by an operator. We performed pivot calibration for the screw driver, which enabled the direct comparison of manually aligned positions of the tool tip to servoed positions. With regards to orientation accuracy, error is measured as the angular distance between the manually aligned and the servoed screw driver shafts.

\section{Results}
\balance
\setcounter{topnumber}{20}
\setcounter{bottomnumber}{20}
\setcounter{totalnumber}{20}
\label{sec:results}

\subsection{ Performance}
\subsubsection{\textbf{EXP-1}}
    \begin{figure*}[t]
        \floatbox[{\capbeside\thisfloatsetup{capbesideposition={right,top},capbesidewidth=4cm}}]{figure}[\FBwidth]
        {\caption{Scatter plot projections comparing the visual servoed and the manually aligned final positions for the goal at the center of the cutting blade at the velocity of 6000~ec/s.}\label{fig:blade_scatter_6000}}
        {\includegraphics[width=11cm]{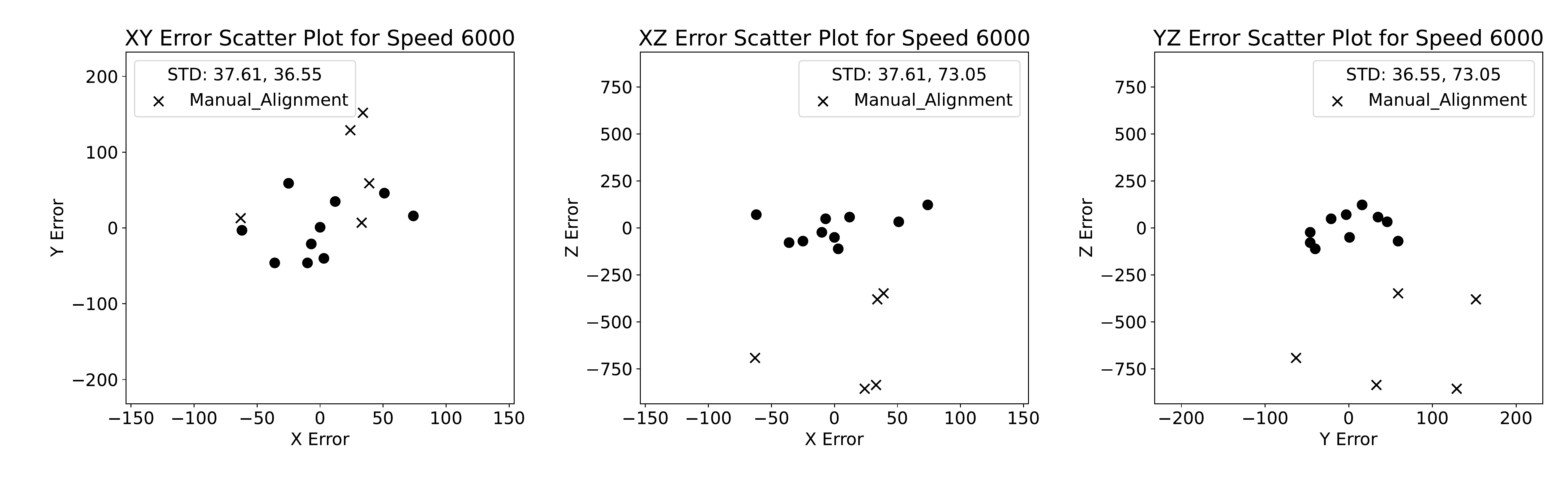}}
   \end{figure*}
   \begin{figure*}[t]
        \floatbox[{\capbeside\thisfloatsetup{capbesideposition={right,top},capbesidewidth=4cm}}]{figure}[\FBwidth]
        {\caption{Scatter plot projections comparing the visual servoed and the manually aligned final positions for the goal at the homing cone vertex at the velocity of 6000~ec/s.}\label{fig:cone_scatter_6000}}
        {\includegraphics[width=11cm]{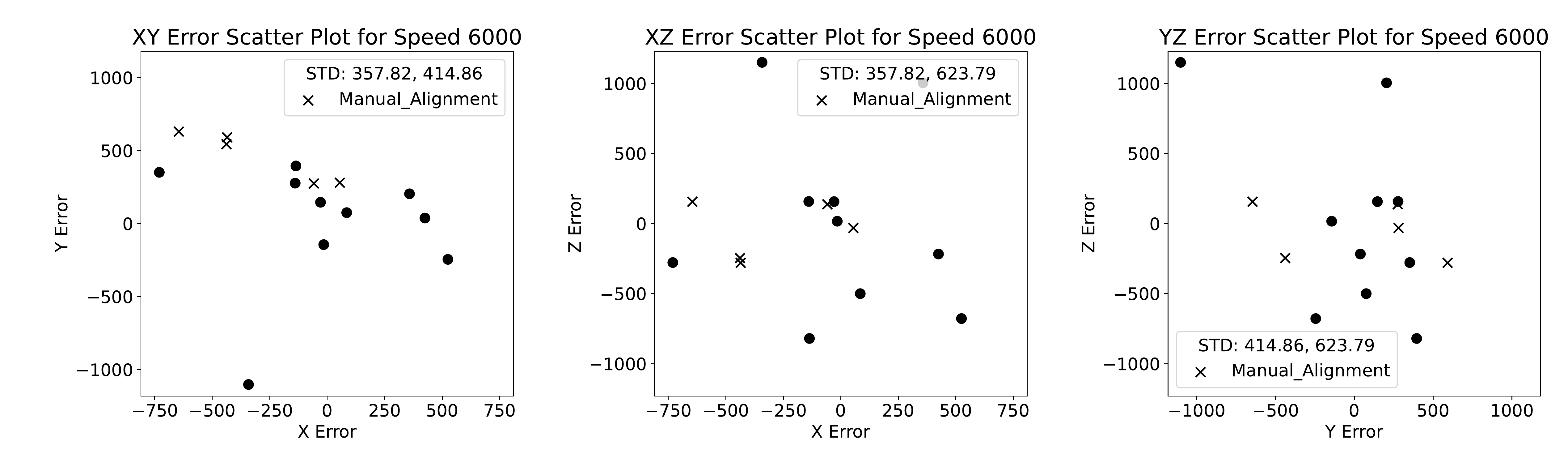}}
    \end{figure*}

    \begin{figure*}[t]
        \floatbox[{\capbeside\thisfloatsetup{capbesideposition={right,top},capbesidewidth=4cm}}]{figure}[\FBwidth]
        {\caption{Scatter plot projections of the visual servoed final positions (and their means) of the tip of the screw driver with respect to the manual alignment position at the three screws.}\label{fig:screws}}
        {\includegraphics[width=11cm]{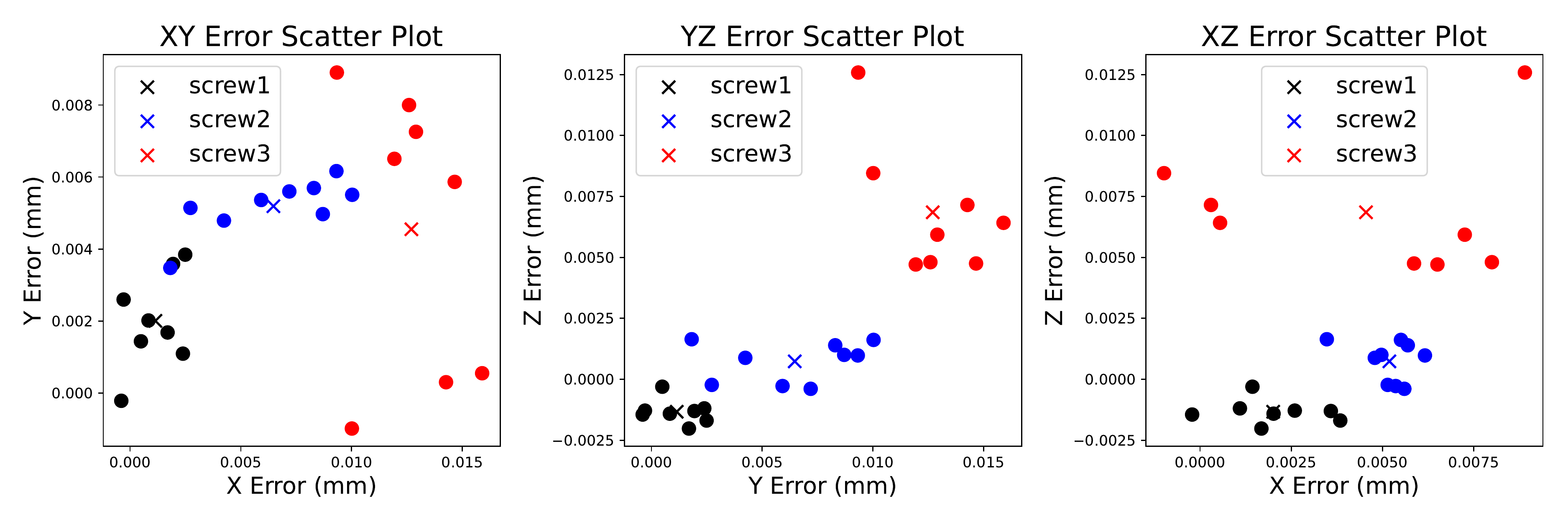}}
    \end{figure*}
    
    % \begin{figure*}[t]
    %     \floatbox[{\capbeside\thisfloatsetup{capbesideposition={right,top},capbesidewidth=4cm}}]{figure}[\FBwidth]
    %     {\caption{scatter plot for foam at 6000 ec/s}\label{fig:foam}}
    %     {\includegraphics[scale=0.5]{foam_scatter_6000.pdf}}
    % \end{figure*}

    For the calibration experiment with the mosquito dissection robot, we ran the full calibration process at different speeds -- 10 times for each speed -- and recorded the calibration results for the three points mentioned in Sec.~\ref{sec:exp1}, and compare this with the ones recorded by manual alignments. To account for the variability in manual alignments across different operators, multiple operators were are asked to perform the alignments and  the mean of these manual calibrations was used in the comparisons. The result is shown in Table~\ref{results_dissection_robot} and Figs.~\ref{fig:blade_scatter_6000}-\ref{fig:cone_scatter_6000} show the detailed scatter plots of the mean-adjusted self-calibration results on the $XY$, $YZ$ and $XZ$ planes.
    \begin{table}[!h]
    
        \begin{adjustbox}{width=\linewidth}
        \caption{(EXP-1) Comparing servoed positions to the mean of manual alignments [mm]. Larger values at the blade center are due to the large differences along the Z axis, and are likely due to humans interpreting the task differently from the YOLOv5 training data.}
        \centering
        \begin{tabular}{|c|c|c|c|c|c|c|c|c|}
    
        % \hline
        % \backslashbox{Type}{Speed} & \multicolumn{2}{|c|}{\centering 2000 ec/sec} & \multicolumn{2}{|c|}{\centering 4000 ec/sec} & \multicolumn{2}{|c|}{\centering 6000 ec/sec} & \multicolumn{2}{|c|}{\centering 8000 ec/sec} \\ \hline
        % & mean & std & mean & std & mean & std & mean & std \\ \hline
        % home & 367.52 & 382.16 & 388.43 & 560.37 & 313.80 & 830.22 & 282.64 & 796.82 \\ \hline
        % blade center & 978.73 & 82.34 & 894.66 & 104.99 & 866.18 & 89.92 & 974.31 & 154.41 \\ \hline
        % tray surface & 35.30 & 18.62 & -84.60 & 137.40 & -34.00 & 15.68 & -25.40 & 35.27 \\ \hline
        \hline
        \backslashbox{Type}{Speed} & \multicolumn{2}{|c|}{\centering 1 mm/sec} & \multicolumn{2}{|c|}{\centering 2 mm/sec} & \multicolumn{2}{|c|}{\centering 3 mm/sec} & \multicolumn{2}{|c|}{\centering 4 mm/sec} \\ \hline
        & mean & std & mean & std & mean & std & mean & std \\ \hline
        home & 0.18376 & 0.19108 & 0.19421 & 0.28019 & 0.15690 & 0.41511 & 0.14132 & 0.39841 \\ \hline
        blade center & 0.48936 & 0.04117 & 0.44733 & 0.05250 & 0.43309 & 0.04496 & 0.48716 & 0.07720 \\ \hline
        tray surface & 0.01765 & 0.00931 & -0.04230 & 0.06870 & -0.01700 & 0.00784 & -0.01270 & 0.01764 \\ \hline
        \end{tabular}
        \label{results_dissection_robot}
        \end{adjustbox}
    \end{table}
    \vspace{-0.2cm}
\subsubsection{\textbf{EXP-2}}
    For the screw-turning experiment, we evaluate the performance of the IBVS method by comparing it to manual alignments obtained through screw calibration. The experiments are conducted at various camera angles. The tool tip frame is pre-determined using pivot calibration. Throughout the experiment, the transformation of the tool tip relative to the robot's base link is monitored. Results are summarized in Table \ref{Screw_Turn_Table}.

    \begin{table}[!h]
    \begin{adjustbox}{width=\textwidth,center}
    
    \caption{Position [m] and orientation [\degree] errors for the robotic screw driver experiment (EXP-2).}
    \centering
    \begin{tabular}{|c|c|c|c|c|c|c|}
    \hline
    \backslashbox{Task}{Angle} & \multicolumn{2}{|c|}{\centering 60\degree} & \multicolumn{2}{|c|}{\centering 90\degree} & \multicolumn{2}{|c|}{\centering 120\degree}  \\ \hline
    & mean & std & mean & std & mean & std \\ \hline
    positional & 0.01585 & 0.00143 & 0.00297 & 0.00110 & 0.00852 & 0.00240 \\ \hline
    angular & 12.73218 & 2.50163 & 5.81744 & 3.04902 & 6.96734 & 1.88652  \\ \hline
    
    \end{tabular}
    \label{Screw_Turn_Table}
    \end{adjustbox}
    \end{table}

    From the results, we can see that the calibration-free visual servoing algorithm can execute the screw-turning task with millimeter-level accuracy and an error of less than 5\degree~when the angle between the two camera views is close to 90\degree. However, the accuracy diminishes as the camera angles are either increased or decreased. In fact, the visual servoing task becomes infeasible when the angle decreases to approximately 30\degree, as the error increases significantly along the center line of the two camera frames. Consequently, even a small detection error can result in a substantial error in Euclidean space.

\captionsetup[figure]{justification=centering}

\subsection{Robot motion analysis} 
In this section, we analyze the robot motions during visual servoing for both experiments, including the trajectory and velocity of the robot.
   \begin{figure*}[!]
      \centering
      \includegraphics[width=1.0\linewidth]{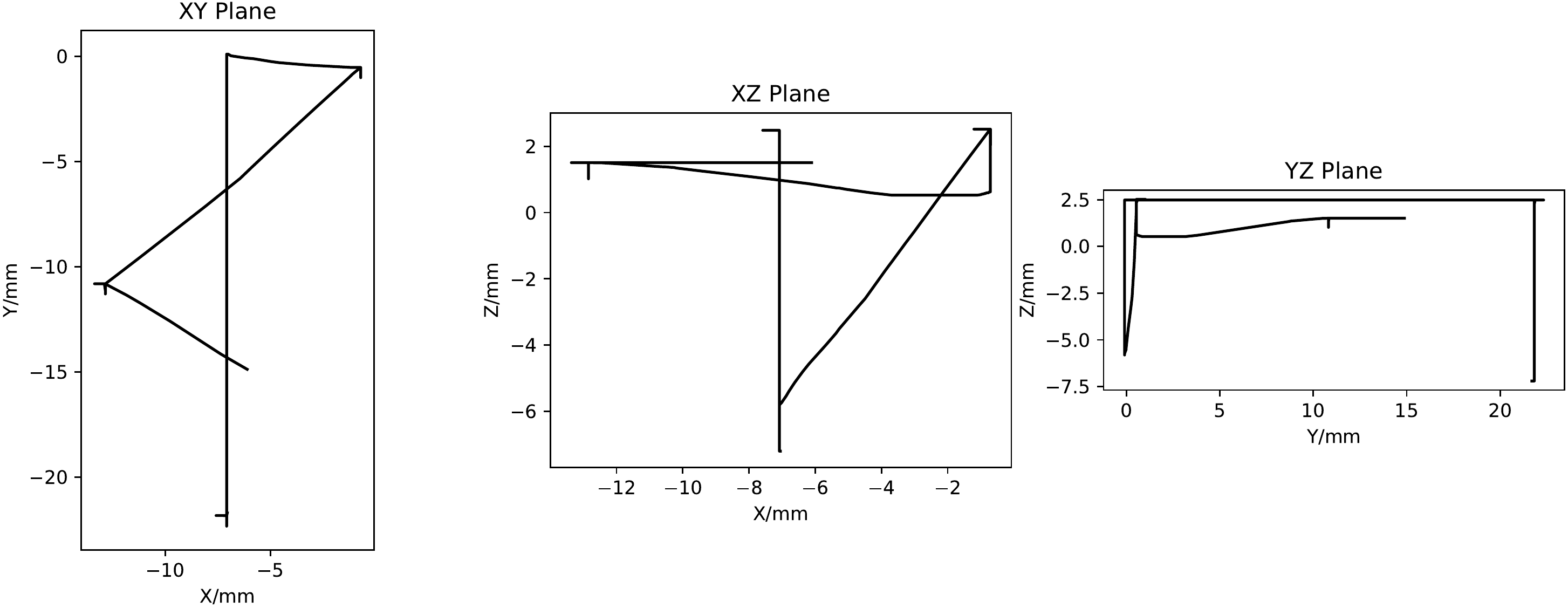}
      \vspace{-30pt}
      \caption{Projections of the robot trajectory recorded while performing full \\ 
      calibration of the robotic mosquito salivary gland extraction system.}
      \label{fig:trajectory}
    \end{figure*}
    
    \begin{figure*}[t]
      \centering
      \includegraphics[width=0.3\linewidth]{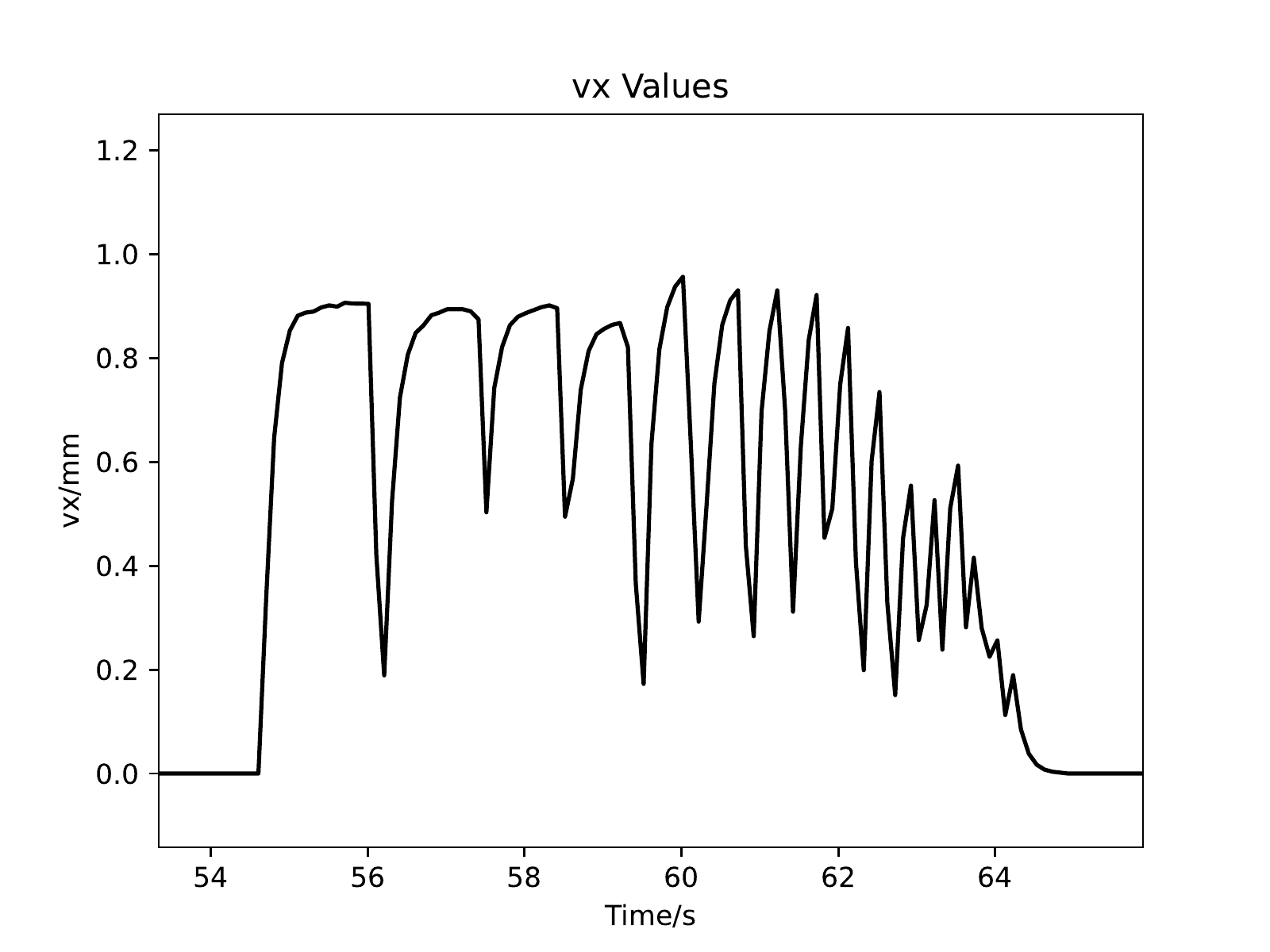}\hfill
      \includegraphics[width=0.3\linewidth]{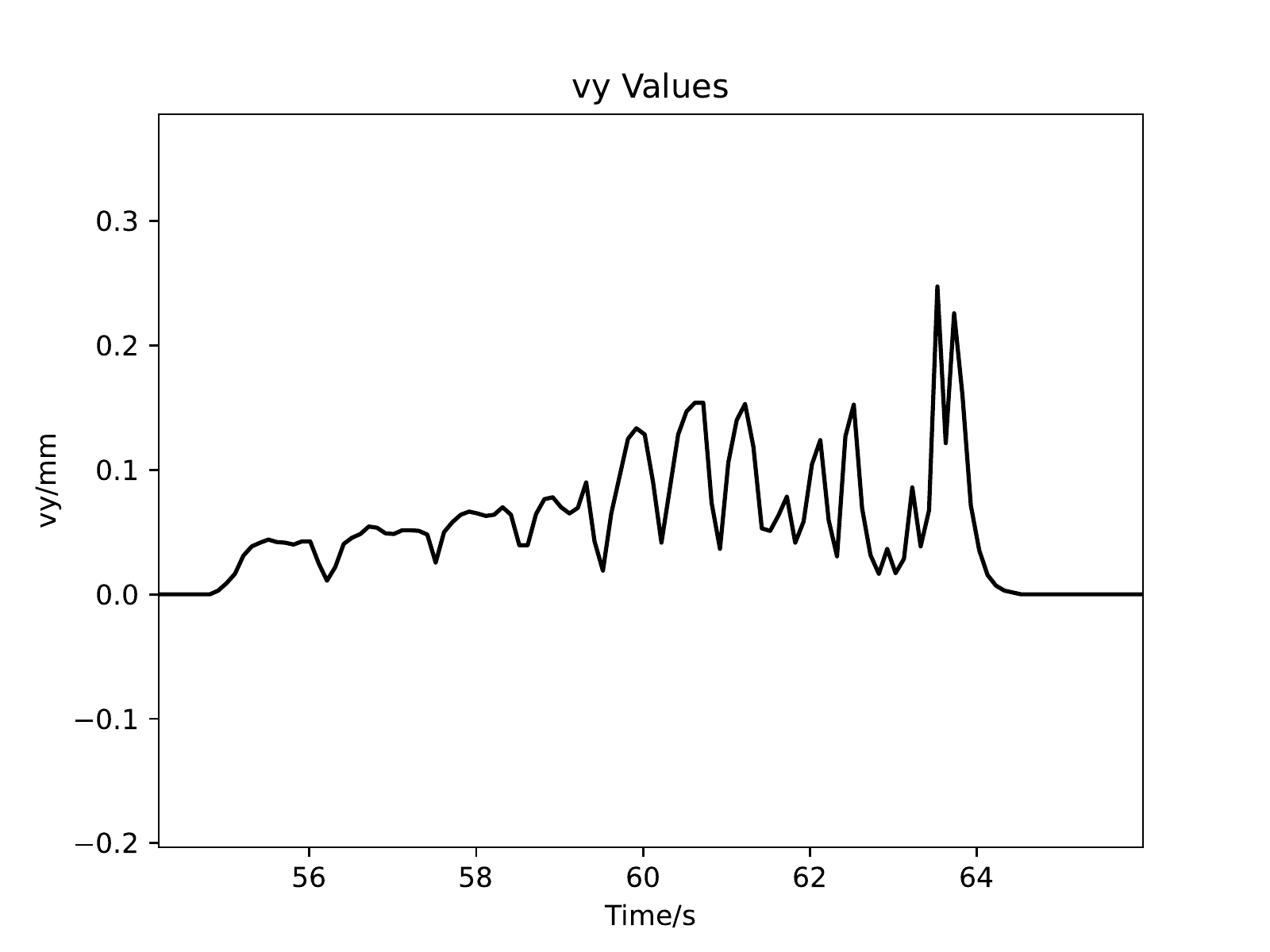}\hfill
      \includegraphics[width=0.3\linewidth]{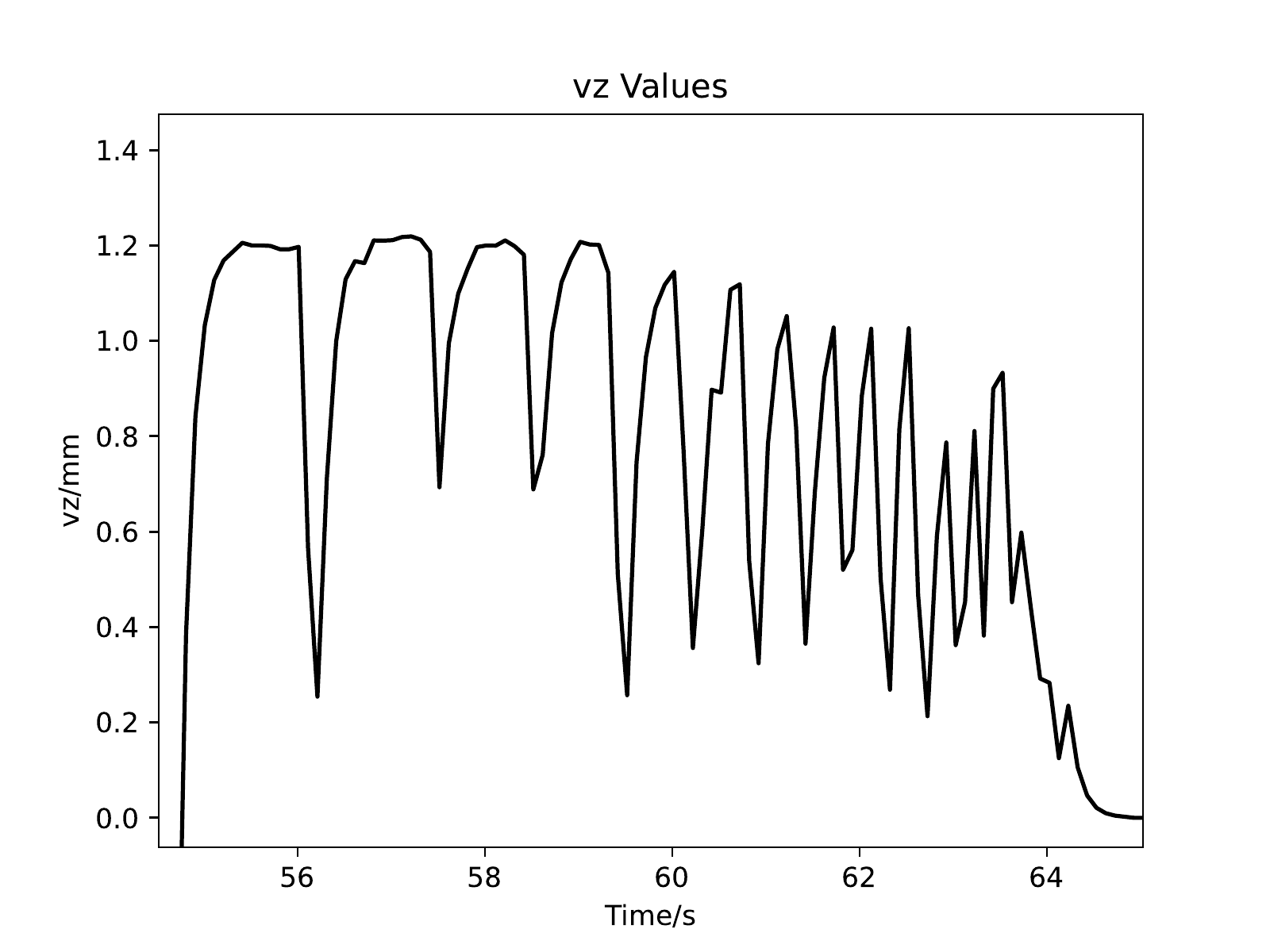}
      \vspace{-0.1cm}
      \caption{Velocity of the robot along the three axes during the visual servoing \\
      segment to approach the center of the cutting blade.}
      \label{fig:blade-vel}
    \end{figure*}
    
\subsubsection{\textbf{EXP-1}}
    Figure \ref{fig:trajectory} shows the projected robot trajectory on the three coordinate planes during full calibration. The short axis-aligned lines at the vertices of the line segments indicate when the robot reinitializes the image Jacobian by performing finite difference approximation. Each line segment indicates a continuous motion of the robot (a point servo to a target feature). These segments are curved because the Jacobian is not constant during the robot motion due to factors like camera distortions, and the direction proposed by the visual servo controller will change slightly from iteration to iteration.
    
    Figure \ref{fig:blade-vel} shows the velocity of the robot during servoing towards the center of the cutting blade. Velocity drops are an artifact of the Galil motion controller hardware (Galil Motion Control Inc., Rocklin, CA, USA) used by the system for driving the Cartesian stage. The controller doesn't allow the preemption of motion commands without first stopping the ongoing motion, which results in a sudden drop of velocity when the Jacobian has an update that affects that particular joint.
    
    \begin{figure*}[!]
          \centering
          \includegraphics[width=0.7\linewidth]{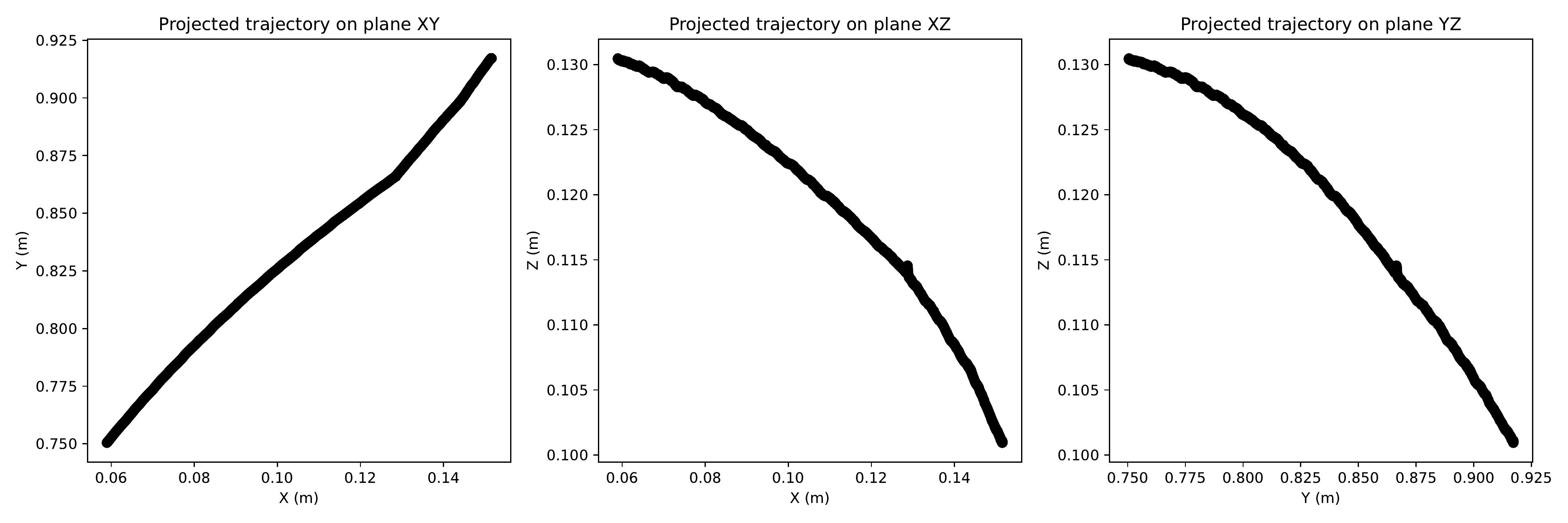}
          \caption{Projections of the robot trajectory when performing position servoing to approach a screw.}
          \label{fig:2D_trajectories_ur10}
          \vspace{-6pt}
    \end{figure*}

    % \begin{figure*}%[htbp]
    %       \centering
    %       \subfloat[\centering rotational servo]{\includegraphics[width=0.5\linewidth]{trajectory_aligning.pdf}}%
    %       \subfloat[\centering positional servo]{\includegraphics[width=0.5\linewidth]{trajectory_position.pdf}\ }%
    %       \vspace{1cm}
    %       \caption{3D trajectories for visual servo of the UR10}%
    %       \label{fig:trajectory_position_and_ori}%
    %       \vspace{-20pt}
    %     \end{figure*}
        
    % \begin{figure*}%[htbp]
    %       \centering
    %       \subfloat[\centering initializing position J]{\includegraphics[width=0.5\linewidth]{trajectory_initialize_J.pdf} }%
    %       \subfloat[\centering initializing rotation J]{\includegraphics[width=0.5\linewidth]{trajectory_initialize_J_ori.pdf} }%
    %       \vspace{1cm}
    %       \caption{3D trajectories for initializing Jacobians of the UR10}%
    %       \label{fig:position_and_orientation_J}%
    %       \vspace{-20pt}
    % \end{figure*}
    
\subsubsection{\textbf{EXP-2}}    
        For the screw alignment task, Figure \ref{fig:2D_trajectories_ur10} illustrates the projected robot trajectory on three coordinate planes during the position servoing part of a visual servo task. As one can observe, the trajectory of the robot on the XY plane contains three line segments, corresponding to three incremental steps in that servo. Each line segment represents a continuous motion of the robot along the way to the target feature. These segments are curved because the Jacobian is not constant during the robot motion due to factors such as camera distortions, causing the direction proposed by the visual servo action to change slightly from iteration to iteration.

\section{Discussion and Conclusions}
\label{sec:conclusions}

The presented image based visual servoing method was developed out of necessity to address the problems that we encountered while calibrating the hardware components of an automated mosquito salivary gland extraction system. The compact size of the device and the high magnification of its cameras prevented us from performing reliable camera calibrations and 3D hand-eye calibration. In addition to this, the relative movements of hardware components in response to vibration and inadvertent touching coupled with the tight tolerances required for the hardware to work correctly mean that calibration has to be done frequently, potentially multiple times a day. The need for manual calibration was a real obstacle in the way of making the system usable in large scale vaccine production.

The implementation of the automated calibration method on the device required only slight changes in the hardware, all of which involved the addition of visual landmarks and markers to the apparatus that facilitate robust image feature detection. The system was already equipped with an array of cameras and a GPU for visually guided mosquito dissection, thus there was no need for expensive upgrades.

In our evaluations, the IBVS-based calibration method has an accuracy comparable to manual calibration. Since we don't have a ground truth for feature locations, the confirmation that our automated calibration results compare favorably to manual calibrations done by multiple human operators was enough to validate the calibration system, which removes one of the outstanding obstacles from the laboratory use of the device.

Furthermore, we extended the method to 5-DoF with the addition of 2 rotational degrees-of-freedom and tested it in a macro-scale task, the fastening of screws using a UR10 robot arm. Our results indicate that this control method is robust and accurate for the subset of tasks where the visual servoing problem can be formulated using relationships between simple 2D image features on multiple views.

We intend to explore further applications for the method in the future, especially in challenging scenarios, such as remote manipulation in hazardous or extreme environments.

\addtolength{\textheight}{-5cm}   % This command serves to balance the column lengths

%%%%%%%%%%%%%%%%%%%%%%%%%%%%%%%%%%%%%%%%%%%%%%%%%%%%%%%%%%%%%%%%%%%%%%%%%%%%%%%%

%%%%%%%%%%%%%%%%%%%%%%%%%%%%%%%%%%%%%%%%%%%%%%%%%%%%%%%%%%%%%%%%%%%%%%%%%%%%%%%%

%%%%%%%%%%%%%%%%%%%%%%%%%%%%%%%%%%%%%%%%%%%%%%%%%%%%%%%%%%%%%%%%%%%%%%%%%%%%%%%%
% \clearpage
% \newpage
%\section*{Acknowledgements}

\bibliographystyle{IEEEtran}   
\bibliography{main}

\end{document}